\documentclass[conference]{IEEEtran}
\IEEEoverridecommandlockouts
\usepackage{cite}
\usepackage{amsmath,amssymb,amsfonts}
\usepackage{algorithmic}
\usepackage{graphicx}
\usepackage{textcomp}
\usepackage{xcolor}
\usepackage{url}

\def\BibTeX{{\rm B\kern-.05em{\sc i\kern-.025em b}\kern-.08em
    T\kern-.1667em\lower.7ex\hbox{E}\kern-.125emX}}
\begin{document}

\title{MedNet and $\mathbf{U}^2$-ViT: A Nested Vision Transformer Diffusion Framework for Diverse Medical Images Generation}

\author{\IEEEauthorblockN{1\textsuperscript{st} Lakshmikar R. Polamreddy}
\IEEEauthorblockA{
\textit{Yeshiva University}\\
New York City, USA \\
lpolamre@mail.yu.edu}
\and
\IEEEauthorblockN{2\textsuperscript{nd} Jialu Li}
\IEEEauthorblockA{
\textit{Boston University}\\
Boston, USA \\
jli21@bu.edu}
\and
\IEEEauthorblockN{3\textsuperscript{rd} Youshan Zhang*}
\IEEEauthorblockA{
\textit{Chuzhou University}\\
Chuzhou, China \\
youshan\_zhang@chzu.edu.cn}
% \and
% \and
% \and
}

\maketitle

\begin{abstract}
The scarcity of diverse, high-quality medical images remains a significant bottleneck in advancing AI-driven medical diagnostics and treatment planning. Traditional datasets often suffer from class imbalances, limited disease coverage, and insufficient variations in imaging conditions — challenges that hinder the development of robust, generalizable models. To address this, we introduce MedNet, a large-scale, multi-modal text-to-image dataset spanning 61 disease categories with 160 classes, covering both humans and animals. We built this with images from trusted public repositories, ensuring data diversity, quality, and ethical compliance. Additionally, we propose $\mathbf{U}^2$-ViT, a novel diffusion-based text-to-image generation architecture arranging Vision Transformers (ViTs) in a nested U-Net structure to enhance the generation of diverse, and high-fidelity medical images. Through extensive experiments, we demonstrate that our model achieves the best Fréchet Inception Distance (FID) score compared to state-of-the-art models. The source code is available at \url{https://github.com/LakshmikarPolamreddy/MedNet-U2-ViT}.\footnote{*Corresponding author; this research was funded by the Research Project of Chuzhou University (Grant No. 2025qd36).}
\end{abstract}


\section{Introduction}
Medical image generation represents a critical frontier in healthcare technology, with far-reaching implications for clinical practice, research, and education. The ability to synthesize realistic medical images offers solutions to persistent challenges in this field: limited availability of annotated data, privacy concerns with patient information, and the need for diverse training examples across rare conditions and demographic groups. Generated medical images can augment training datasets for diagnostic AI systems, facilitate medical education without patient data exposure, and enable visualization of disease progression to improve treatment planning. Furthermore, text-guided medical image synthesis creates an interface for clinicians to generate visual representations of textual findings, enhancing communication and understanding of medical conditions. 

Recent advances in diffusion models have significantly transformed the medical imaging landscape. Researchers have applied these generative approaches across various domains, from reconstruction to synthesis and text-to-image generation. In reconstruction, diffusion-based methods have enabled high-quality image recovery from limited measurements, reducing radiation exposure and acquisition times while maintaining diagnostic accuracy. For image synthesis, researchers have developed specialized frameworks for generating realistic medical images across modalities including histopathology, echocardiography, mammography, and various forms of radiological imaging. More recently, text-to-image synthesis models have begun translating radiological reports into corresponding visual representations, creating a bridge between textual and visual medical information. Despite these advancements, existing approaches remain constrained by narrow scope-typically focusing on single modalities or limited disease categories (usually 1-3 classes). This specialization, while effective for specific applications, limits the development of generalizable frameworks that could serve broader clinical and research needs. Additionally, many current approaches rely on private or restricted datasets, limiting reproducibility and widespread adoption.

\begin{figure}
    \centering
    \includegraphics[width=0.95\linewidth]{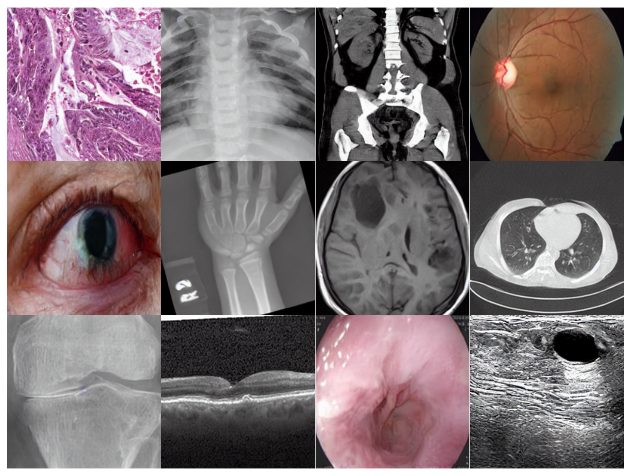}
    \caption{$\mathbf{U}^2$-ViT generated images of 256$\times$256 resolution using MedNet.}
    \label{fig:u2vit_gen}
\end{figure}

To address these limitations, we introduce MedNet, a large-scale, multi-class medical image dataset specifically designed to advance AI research in medical diagnostics. This dataset encompasses 241,598 images across 160 classes organized into 61 high-level disease categories. A key advantage of our work is its comprehensive scope, covering not only human diseases but also conditions affecting various animals including dogs, hens, cattle, and others-enabling research applications in both human and veterinary medicine. In addition, we develop $\mathbf{U}^2$-ViT model, a comprehensive text-to-image synthesis model capable of generating realistic medical images across all 160 distinct classes. Moreover, we concatenate the outputs from each of bottleneck, and decoder layers for better image synthesis. Fig.~\ref{fig:u2vit_gen} demonstrates the quality of medical images generated by $\mathbf{U}^2$-ViT model.

\section{Related Work}\label{sec:rel}
Recent advances in diffusion models have significantly improved medical image synthesis, reconstruction, augmentation, and text-to-image generation~\cite{polamreddy2025deficit}.

\textbf{Medical Image Synthesis using Diffusion Models:}
Diffusion models have been widely adopted for medical image generation across diverse modalities and applications. Existing works explored echocardiography video synthesis~\cite{van2024training}, data augmentation for canine cardiomegaly detection~\cite{zhang2025confident}, high-resolution cellular image generation~\cite{cechnicka2024urcdm}, disease progression modeling~\cite{puglisi2024enhancing}, mammography synthesis~\cite{na2024radiomicsfill}, ultrasound generation~\cite{dominguez2024diffusion}, 
%breast cancer analysis~\cite{salem2024deep}, 
anatomical structure modeling~\cite{sinha2024trind}, multi-modal synthesis~\cite{zhan2024medm2g,jiang2023cola,li2023ddmm}, severity-aware generation~\cite{takezaki2024ordinal}, tumor synthesis~\cite{chen2024towards}, %data augmentation~\cite{zhong2024meddiffusion,ye2023synthetic,sagers2022improving}, 
CT and MRI synthesis~\cite{txurio2023diffusion,abd2024sdlm,pinaya2022brain}, %2D/3D medical image generation~\cite{pan20232d,dorjsembe2022three,khader2023denoising}, 
temporal image generation~\cite{kim2022diffusion}, and semantic diffusion for colonoscopy synthesis~\cite{du2023arsdm}. Collectively, these approaches address data scarcity, enable privacy-preserving data sharing, improve diagnostic model robustness through synthetic data augmentation, and facilitate multi-modal medical image analysis.

\textbf{Text-to-Image Synthesis in Medical Imaging:}
Recent text-guided diffusion models have extended medical image synthesis by leveraging clinical text descriptions. Representative methods include Chest-Diffusion for report-to-X-ray generation~\cite{huang2024chest}, MediSyn for multi-modal text-guided synthesis~\cite{cho2025medisyngeneralisttextguidedlatent}, DiffBoost for text-guided segmentation~\cite{zhang2024diffboost}, Fair Text-to-Medical Image Diffusion for bias-aware image generation~\cite{han2024fair}, and Text-to-Image generation of medical images using LLCM~\cite{polamreddy2025leapfrog}. These methods improve report visualization, support radiology education, enhance segmentation with textual guidance, and promote fairer medical AI systems.

Despite these advances, existing methods typically focus on a single disease or a small number of medical conditions. In contrast, our work presents a unified text-to-image diffusion framework capable of generating realistic medical images across \textbf{160 disease classes}, substantially improving scalability and clinical diversity.

\section{Dataset}

\begin{figure*}
    \centering
    \includegraphics[width=\linewidth]{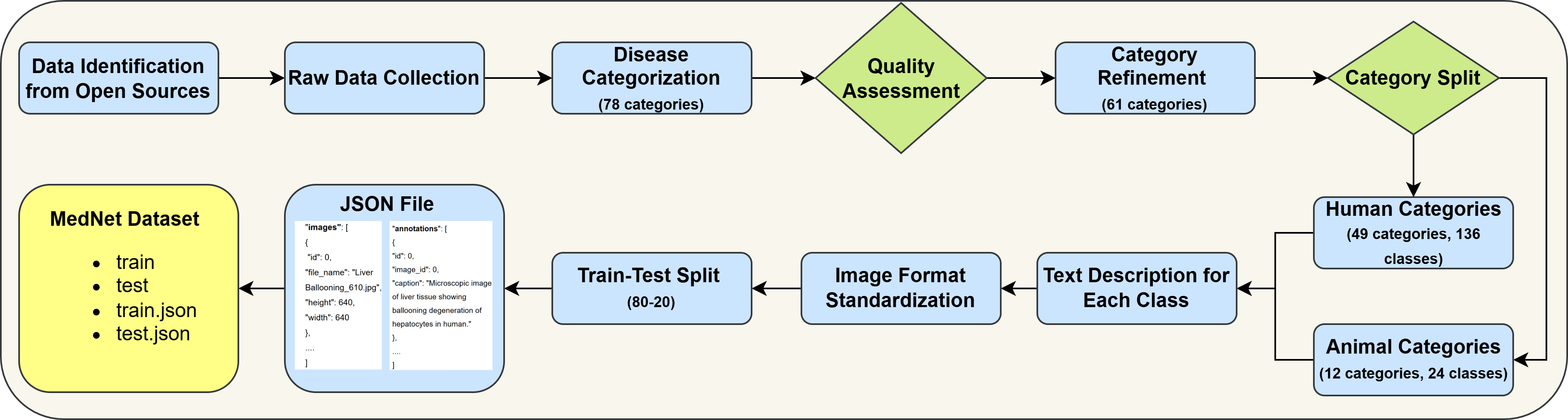}
    \caption{Overview of the MedNet dataset construction pipeline. Images are collected from multiple public repositories and specialized medical datasets, followed by data cleaning, format standardization, and organization into human and animal disease domains across multiple imaging modalities.}
    \label{fig:mednet_pipeline}
\end{figure*}

We introduce \textbf{MedNet}, a large-scale multi-modal medical image dataset designed for medical image synthesis and diagnostic modeling (Figure~\ref{fig:mednet_pipeline}). Images were collected from publicly available repositories~\cite{kaggle_website,github_website,roboflow_website} and specialized medical datasets, including HAM10000~\cite{DVN/DBW86T_2018} and the Knee X-ray Osteoporosis Database~\cite{Majeed2021}. Only datasets with publicly available usage rights and clear licensing terms were included.

MedNet comprises \textbf{241,598 images} spanning \textbf{78 disease categories}, refined into \textbf{61 categories} and \textbf{160 disease classes}. The dataset consists of \textbf{136 human disease classes} across \textbf{49 disease categories} and \textbf{24 animal disease classes} across \textbf{12 disease categories}. To ensure data quality and consistency, corrupted and duplicate images were removed, all images were converted to PNG format, and an \textbf{80:20} class-balanced train--test split was created, resulting in \textbf{193,163} training and \textbf{48,435} testing images.

MedNet covers \textbf{nine imaging modalities}: X-ray, MRI, CT, Ultrasound, Dermatoscopic, Fundus/Retinal OCT, Histopathological, Endoscopic, and Microscopic imaging. This diversity reflects real-world clinical practice and supports robust cross-modality learning for medical image synthesis and downstream diagnostic tasks. The dataset is organized into two domains: human diseases and animal diseases.

\textbf{Human Diseases.}
The human subset contains \textbf{49 disease categories} and \textbf{136 classes}, covering diverse anatomical regions and clinical conditions, including neurological disorders, cancers, ophthalmic diseases, pulmonary diseases, musculoskeletal disorders, and dermatological conditions. Several diseases are further divided into clinically meaningful subclasses (e.g., Alzheimer's disease severity levels and benign/malignant tumors), enabling fine-grained medical image synthesis.

\textbf{Animal Diseases.}
The animal subset contains \textbf{12 disease categories} and \textbf{24 classes} across multiple species, including cattle, dogs, fish, and poultry. It covers skin, eye, respiratory, musculoskeletal, and infectious diseases, supporting veterinary image synthesis and comparative cross-species medical AI research.

\section{Methods}\label{sec:met}

\textbf{Overall Pipeline for Image Generation. }We generate images by integrating a hybrid convolutional and transformer-based design, leveraging a U-ViT structure to enhance the diffusion model’s ability to generate diverse, high-quality images. The process begins with a real image undergoing a forward diffusion process to produce a noisy image~\cite{ho2020denoising}. The forward diffusion process gradually adds Gaussian noise to an image $\mathbf{x}_0$ over $T$ time steps:

\begin{equation}
q(\mathbf{x}_t | \mathbf{x}_{t-1}) = \mathcal{N}(\mathbf{x}_t; \sqrt{1 - \beta_t} \, \mathbf{x}_{t-1}, \beta_t \mathbf{I}),
\end{equation}

where $\beta_t$ is the variance schedule. The complete forward process is defined as:

\begin{equation}
q(\mathbf{x}_{1:T} | \mathbf{x}_0) = \prod_{t=1}^{T} q(\mathbf{x}_t | \mathbf{x}_{t-1}).
\end{equation}

The reverse diffusion process aims to reconstruct the original image $\mathbf{x}_0$ from noise using a learned model $p_\theta(\mathbf{x}_{t-1} | \mathbf{x}_t)$:

\begin{equation}
p_\theta(\mathbf{x}_{0:T}) = p(\mathbf{x}_T) \prod_{t=1}^{T} p_\theta(\mathbf{x}_{t-1} | \mathbf{x}_t),
\end{equation}

where each $p_\theta(\mathbf{x}_{t-1} | \mathbf{x}_t)$ is parameterized as a Gaussian distribution whose mean is predicted by the model.

The noisy image, along with its conditioned text is passed through an embedding layer that encodes spatial and temporal information. A convolutional layer extracts preliminary features, which are then fed into the $\mathbf{U}^2$-ViT module—a hierarchical structure composed of multiple U-ViT blocks interconnected with skip connections. Each U-ViT block applies Vision Transformer (ViT) layers to capture both global dependencies and local details, improving the model’s understanding of complex patterns. The skip connections ensure low-level features are preserved throughout the network, aiding in reconstructing fine details. The output is progressively denoised through a decoder resembling a U-Net, equipped with ViT blocks at each level, culminating in a bottleneck of enhanced transformer layers to handle the most abstract representations. Finally, a sampling module reconstructs the generated image from the predicted noise, completing the reverse diffusion process. This combination of convolutional encoders, nested transformers, and skip connections enables the model to retain fine-grained details while capturing large-scale structures, making it particularly effective for generating diverse and realistic samples.

\textbf{Core Model Architecture. } The core of the architecture is the $\mathbf{U}^2$-ViT module, composed of five main U-ViT blocks - Two encoder U-ViT blocks, One bottleneck U-ViT block, and Two decoder U-ViT blocks. Each U-ViT block captures multi-scale information through transformer layers while retaining low-level details via skip connections. Each block contains:
\begin{itemize}
    \item \textbf{8 encoder ViT blocks:} Responsible for progressively encoding the input features and capturing global patterns.
    \item \textbf{1 bottleneck ViT block:} Positioned at the center of the block to capture the most abstract feature representations.
    \item \textbf{8 decoder ViT blocks:} Symmetrically reverse the encoding process, refining features and reconstructing spatial details of the image.
\end{itemize}

The output of the $\mathbf{U}^2$-ViT is formed by concatenating the outputs from the bottleneck U-ViT block and the two decoder U-ViT blocks. This fusion combines abstract semantic information with fine-grained spatial details, facilitating the generation of high-fidelity images with rich contextual features.

\begin{equation}
\mathbf{F}_{\text{$\mathbf{U}^2$-ViT}} = \text{Concat}(\mathbf{F}_{\text{bottleneck}}, \mathbf{F}_{\text{dec1}}, \mathbf{F}_{\text{dec2}})
\end{equation}

This nested structure enables a hierarchical learning process that effectively balances fine detail reconstruction with high-level semantic understanding.

\textbf{Training Methodology. }To train the Nested U-ViT efficiently and ensure high-quality generation, we employ a two-stage training strategy:
\begin{itemize}
    \item \textbf{Stage 1: Pretraining Base U-ViT. }We begin by training a single standalone U-ViT model for 300\text{K} iterations. This model learns to predict the noise added during the forward diffusion process by minimizing the mean squared error (MSE) loss~\cite{ho2020denoising}:
    \begin{equation}
    \mathcal{L}_{\text{MSE}} = \mathbb{E}_{t, \mathbf{x}_0, \boldsymbol{\epsilon}} \left[ \left\| \boldsymbol{\epsilon} - \boldsymbol{\epsilon}_\theta(\mathbf{x}_t, t) \right\|^2 \right].
    \end{equation}

    \item \textbf{Stage 2: Fine-tuning $\mathbf{U}^2$-ViT. }We then embed the trained U-ViT parameters into the larger $\mathbf{U}^2$-ViT structure and freeze these blocks. This initialization significantly accelerates training and improves stability. The remaining components of $\mathbf{U}^2$-ViT are fine-tuned for 30\text{K} additional iterations. The loss is now optimized with respect to the parameters of the nested structure.
    \begin{equation}
    \min_{\theta'} \, \mathbb{E} \left[ \left\| \boldsymbol{\epsilon} - \boldsymbol{\epsilon}_{\theta'}(\mathbf{x}_t, t; E_{\text{base}}) \right\|^2 \right],
    \end{equation}
    where \( E_{\text{base}} \) denotes the frozen encoder-decoder backbone from Stage 1.
\end{itemize}
This two-stage strategy ensures faster convergence, reduces computational cost, and improves overall generation quality by leveraging the strengths of both pretrained and hierarchical designs. % Algorithm~\ref{alg:nested_u2vit} outlines the complete workflow of our proposed text-to-image architecture for medical images generation.

\begin{figure*}
    \centering
    \includegraphics[width= 0.8\linewidth]{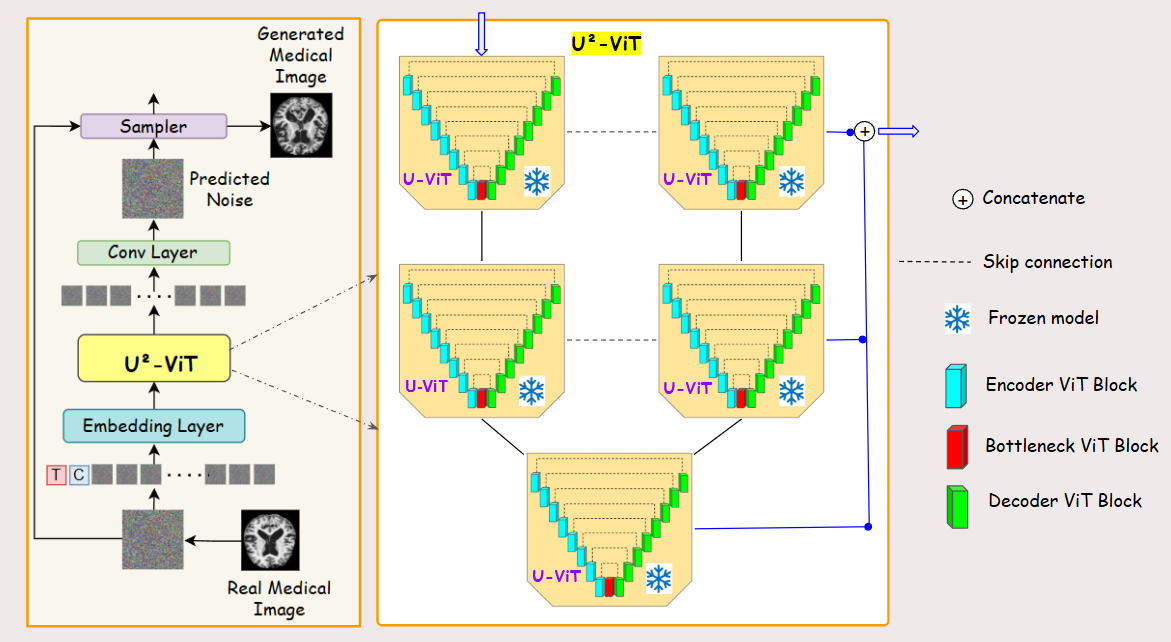}
    \caption{
Overview of the proposed $\mathbf{U}^2$-ViT diffusion architecture for text-guided medical image generation.}
    \label{fig:model}
\end{figure*}

\section{Experiments and Results}\label{sec:exp}

\subsection{Implementation details}
We trained U-ViT for 300K iterations and $\mathbf{U}^2$-ViT model for 30K iterations with a batch size of 256, utilizing $2\times$ A100 GPUs. We utilized the AdamW optimizer with a learning rate of \(2 \times 10^{-4}\), weight decay of 0.03, and $\beta$ parameters set to (0.9, 0.9). We employed the DDPM scheduler~\cite{ho2020denoising}, a pre-trained image autoencoder~\cite{kingma2013auto}, and DPM-Solver~\cite{lu2022fast}. To ensure a fair comparison, all state-of-the-art models are trained using the same computational resources and similar hyperparameter settings.

\subsection{Results}
We conduct extensive experiments on the MedNet dataset to evaluate the performance of the proposed $\mathbf{U}^2$-ViT model and compare it with several state-of-the-art (SOTA) text-to-image generation methods. The evaluation includes both qualitative and quantitative analyses to assess the model’s ability to generate medically meaningful, high-fidelity images. These experiments demonstrate the effectiveness of our architecture in capturing fine-grained anatomical structures and producing realistic medical imagery.

\begin{figure*}
    \centering
    \includegraphics[width= 0.8\linewidth]{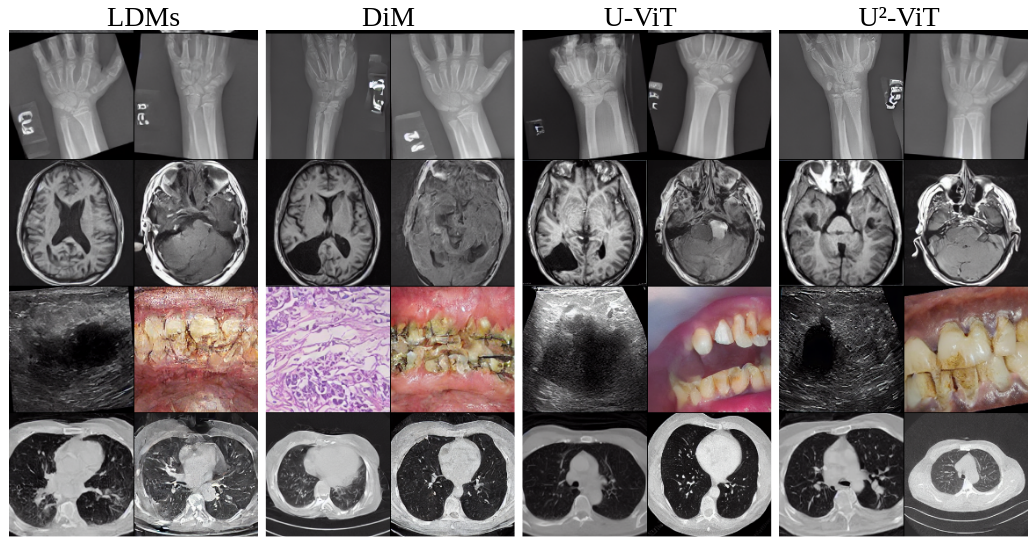}
    \caption{
Qualitative comparison of medical images generated by different text-to-image models on the MedNet dataset. These results demonstrate the ability of $\mathbf{U}^2$-ViT to better capture fine-grained medical details and produce high-fidelity images.}
    \label{fig:comp}
\end{figure*}

\textbf{Qualitative results. }We present the qualitative comparison between the proposed $\mathbf{U}^2$-ViT and existing state-of-the-art (SOTA) text-to-image models on the MedNet dataset. As shown in Figure \ref{fig:comp}, the $\mathbf{U}^2$-ViT model consistently generates more realistic and detailed medical images. One of the most noticeable differences is observed in the generation of medical features, such as bones and teeth. The $\mathbf{U}^2$-ViT model produces sharper and more anatomically accurate representations of these structures, highlighting finer details like the texture and contours of bones and the clarity of teeth. In contrast, the images generated by the other tend to be less defined, with a noticeable lack of fine-grained detail, especially in complex structures like teeth.

\textbf{Quantitative results. }Table~\ref{tab:fid_scores} presents a comparison of Fréchet Inception Distance (FID) scores achieved by $\mathbf{U}^2$-ViT and several state-of-the-art (SOTA) text-to-image generation models on the MedNet dataset. The proposed $\mathbf{U}^2$-ViT significantly outperforms all competing methods, achieving the lowest FID score of 23.04, indicating superior visual quality and fidelity in generated medical images. In contrast, existing diffusion-based models such as DDPM (52.33), LDMs (45.60), and DreamBooth (48.15) yield much higher FID values, reflecting their limitations in capturing the fine-grained anatomical details crucial in medical imagery. Recent models like U-ViT (33.70) and DiM (36.42) show improvements over traditional diffusion models, but still fall short compared to $\mathbf{U}^2$-ViT. The results clearly demonstrate the effectiveness of our dual-stage architecture and the advantages of integrating both U-shaped and Vision Transformer components tailored for medical image generation. However, we still have scope to improve the performance of our model by enhancing the image quality, as demonstrated by reducing the FID score to below 5.

\begin{table*}
    \centering
    \caption{
Quantitative comparison of $\mathbf{U}^2$-ViT with state-of-the-art text-to-image generation models on the MedNet dataset. }
    \begin{tabular}{lllllll}
    \hline
    Model & Params(M) & Year & Venue & FID & Precision & Recall \\
    \hline
    DDPM~\cite{ho2020denoising} & - & 2020 & NeurIPS & 52.33 & 0.71 & 0.48 \\
    LDMs~\cite{rombach2022high} & 400 & 2022 & CVPR & 45.60 & 0.74 & 0.48 \\
    Dreambooth~\cite{ruiz2023dreambooth} & - & 2023 & CVPR & 48.15 & 0.68 & 0.49 \\
    LCMs~\cite{luo2023latent} & - & 2023 & - & 56.24 & 0.64 & 0.44 \\
    U-ViT~\cite{bao2023all} & 58 & 2023 & CVPR &  33.70 & 0.76 & 0.51\\
    DiM~\cite{teng2024dim} & 380 & 2024 & - & 36.42 & 0.75 & 0.49 \\
    $\mathbf{U}^2$-ViT & 292 & 2026 & Healthcom & \textbf{23.04 } & \textbf{0.78} & \textbf{0.52}\\
    \hline
    \end{tabular}%
    \label{tab:fid_scores}
\end{table*}

\subsection{Ablation Study}

We conduct a two-stage ablation study on the MedNet dataset. First, we identify the optimal U-ViT configuration by varying the number of encoder and decoder layers while keeping a single bottleneck layer fixed. Second, using the best U-ViT configuration, we evaluate different hierarchical designs of the proposed $\mathbf{U}^2$-ViT architecture with and without intermediate feature concatenation.

\noindent\textbf{U-ViT Configuration.}
Table~\ref{tab:uvit} reports the performance of different U-ViT depths. Increasing the model depth consistently improves the FID score up to 8 encoder-decoder layers, achieving the best result of \textbf{33.70} with 58M parameters. Deeper models show a slight degradation, indicating diminishing returns. Therefore, we adopt the 8E--1B--8D configuration as the building block for the proposed $\mathbf{U}^2$-ViT.

\begin{table}
    \centering
    \caption{FID scores on the MedNet dataset for different U-ViT configurations. E, B, and D denote encoder, bottleneck, and decoder layers, respectively.}
    \begin{tabular}{lll}
    \hline
    U-ViT Configuration & Params (M) & FID \\
    \hline
    2$\times$E, B, 2$\times$D & 17 & 182.44 \\
    3$\times$E, B, 3$\times$D & 24 & 160.50 \\
    4$\times$E, B, 4$\times$D & 31 & 132.00 \\
    5$\times$E, B, 5$\times$D & 38 & 100.20 \\
    6$\times$E, B, 6$\times$D & 45 & 50.12 \\
    7$\times$E, B, 7$\times$D & 52 & 38.26 \\
    8$\times$E, B, 8$\times$D & 58 & \textbf{33.70} \\
    9$\times$E, B, 9$\times$D & 65 & 35.61 \\
    10$\times$E, B, 10$\times$D & 72 & 37.24 \\
    11$\times$E, B, 11$\times$D & 79 & 40.26 \\
    12$\times$E, B, 12$\times$D & 86 & 42.05 \\
    \hline
    \end{tabular}
    \label{tab:uvit}
\end{table}

\noindent\textbf{$\mathbf{U}^2$-ViT Configuration.}
Using the optimal U-ViT block, we evaluate several hierarchical $\mathbf{U}^2$-ViT designs (Table~\ref{tab:u2vit_config}). Without feature concatenation, the 2-layer model achieves the best FID of \textbf{32.15}, outperforming both shallower and deeper variants. Incorporating intermediate bottleneck and decoder feature concatenation further improves all configurations, with the 2-layer model achieving the best overall FID of \textbf{23.04}. These results demonstrate that hierarchical feature reuse is more beneficial than simply increasing network depth. Consequently, the 2-layer concatenated $\mathbf{U}^2$-ViT is selected as the final architecture.

\begin{table}
    \centering
    \caption{FID scores on the MedNet dataset for different $\mathbf{U}^2$-ViT configurations. $FID_{w/o-C}$ and $FID_C$ denote results without and with intermediate feature concatenation, respectively.}
    \begin{tabular}{lll}
    \hline
    $\mathbf{U}^2$-ViT Configuration & $FID_{w/o-C}$ & $FID_C$ \\
    \hline
    Linear stack of 3 U-ViTs & 38.22 & -- \\
    1$\times$E, B, 1$\times$D & 35.99 & 34.20 \\
    2$\times$E, B, 2$\times$D & 32.15 & \textbf{23.04} \\
    3$\times$E, B, 3$\times$D & 33.91 & 28.16 \\
    4$\times$E, B, 4$\times$D & 35.45 & 29.20 \\
    \hline
    \end{tabular}
    \label{tab:u2vit_config}
\end{table}

\section{Conclusion}\label{sec:con}
We create MedNet, a comprehensive, large-scale, multi-modal medical image dataset designed to address the limitations of existing medical imaging resources. By encompassing 241,598 images across 160 classes and 61 disease categories, MedNet ensures data diversity and quality, supporting advancements in both human and veterinary medical applications. Furthermore, we propose $\mathbf{U}^2$-ViT model, a diffusion-based text-to-image model designed to enhance the generation of diverse medical images. Our experiments demonstrate the superior performance of our model, with significant improvements in image quality compared to the existing SOTA models. In future work, we plan to expand the MedNet dataset to include additional disease categories.
\bibliographystyle{splncs04}
\bibliography{mybibliography}
\end{document}